\documentclass[twocolumn]{article}

\usepackage{fullpage,times,amsmath,epsfig}
\usepackage{luca}
\usepackage{hyperref}
\usepackage[linesnumbered, vlined]{algorithm2e}
\usepackage{algpseudocode}
\usepackage{times}
\usepackage{graphicx}
\usepackage{verbatim}
\usepackage{caption}
\usepackage{color}
\usepackage{pbox}
\usepackage{comment}

\usepackage{amsmath}
\title{Learning From Graph Neighborhoods Using LSTMs}

\author{
  Rakshit Agrawal\thanks{The authors prefer to be listed in alphabetical order.} \and Luca de Alfaro \and Vassilis Polychronopoulos\\
  ragrawa1@ucsc.edu, luca@ucsc.edu, vassilis@cs.ucsc.edu\\
  Computer Science Department, University of California, Santa Cruz\\[1ex]
  Technical Report UCSC-SOE-16-17\\
  School of Engineering, UC Santa Cruz
}
\date{November 18, 2016}

\def\loss{{\cal L}}
\def\concat{{\stackrel{\frown}{}}}

\begin{document}
\maketitle

\begin{abstract} 
Many prediction problems can be phrased as inferences over local neighborhoods of graphs.  The graph represents the interaction between entities, and the neighborhood of each entity contains information that allows the inferences or predictions. We present an approach for applying machine learning directly to such graph neighborhoods, yielding predicitons for graph nodes on the basis of the structure of their local neighborhood and the features of the nodes in it.
Our approach allows predictions to be learned directly from examples, bypassing the step of creating and tuning an inference model or summarizing the neighborhoods via a fixed set of hand-crafted features. The approach is based on a multi-level architecture built from Long Short-Term Memory neural nets (LSTMs); the LSTMs learn how to summarize the neighborhood from data. We demonstrate the effectiveness of the proposed technique on a synthetic example and on real-world data related to crowdsourced grading, Bitcoin transactions, and Wikipedia edit reversions.
\end{abstract}

%
\section{Introduction}

Many prediction problems can be naturally phrased as inference problems over the local neighborhood of a graph. 
Consider, for instance, crowdsourced grading. 
We can construct a (bipartite) graph consisting of items and graders, where edges connect items to users who graded them, and are labeled with the grade assigned.
To infer the grade for an item, we can look at the graph involving the adjacent nodes: this graph, known as the 1-neighborhood, consists of the people who graded the item and of the grades they assigned.
If we wish to be more sophisticated, and try to determine which of these people are good graders, we could look also at the work performed by these people, expanding our analysis outwards to the 2- or 3-neighborhood of each item. 

For another example, consider the problem of predicting which bitcoin addresses will spend their deposited funds in the near future.
Bitcoins are held in ``addresses''; these addresses can participate in transactions where they send or receive bitcoins.
To predict which addresses are likely to spend their bitcoin in the near future, it is natural to build a graph of addresses and transactions, and consider neighborhoods of each address.
The neighborhood contains information on where the bitcoins came from, and on what happened to bitcoins at the interacting addresses, which (as we will show) can help predict whether the coins will be transacted soon.

For a third example, consider the problem of predicting user behavior on Wikipedia. 
Users interact by collaboratively editing articles, and we are interested in predicting which users will have their work reverted.
We can build a graph with users as nodes, and interactions as edges: an interaction occurs when two users edit the same article in short succession, and one either keeps, or undoes, the work of the other. 
The 1-neighborhood of a user will tell us how often that user's work has been kept or reverted. 
Again, we can consider larger neighborhoods to gather information not only on the user, but on the people she interacted with, trying to determine whether they are good contributors, how experienced they are, whether they are involved in any disputes, and so forth.

In this paper, we show how to solve these problems by applying machine learning, using an architecture based on multi-level {\em Long Short-Term Memory\/} (LSTM) neural nets \cite{hochreiter_long_1997,gers_lstm_2001,graves_supervised_2012}, with each LSTM level processing one ``degree of separation'' in the neighborhood.

The challenge of applying machine learning to graph neighborhoods lies in the fact that many common machine learning methods, from neural nets \cite{hopfield_neural_1982} to support vector machines (SVMs) \cite{cortes_support-vector_1995}, are set up to handle fixed-length vectors of features as input. 
As a graph neighborhood is variable in size and topology, it is necessary to summarize the neighborhood into a fixed number of features to use in learning. 
Some machine learning methods, such as logistic regression \cite{bishop_pattern_2007}, can accept a potentially unbounded number of inputs, but every input has its own index or name, and it is not obvious how to map the local topology of a graph into such fixed naming scheme in a way that preserves the structure, or the useful information.

Machine-learning methods that can learn from sequences, such as LSTMs \ or recurrent neural nets \cite{williams_gradient-based_1995,hochreiter_gradient_2001}, offer more power. 
It is possible to traverse the local neighborhood of a node in a graph in some order (pre-, post-, or in-order), and encode the neighborhood in a sequence of features complete with markers to denote edge traversals, and then feed this sequence to an LSTM. 
We experimented with this approach, but we did not obtain any useful results: the LSTMs were unable to learn anything useful from a flattened presentation of the graph neighborhood.

We propose a learning architecture based on the use of multiple levels of LSTMs. 
Our architecture performs predictions for one ``target'' graph node at a time. 
First, the graph is unfolded from the target node, yielding a tree with the target node as its root at level 0, its neighbors as level-1 children, its neighbors' neighbors as level-2 children, and so forth, up to a desired depth $D$.
At each tree node $v$ of level $0 \leq d < D$, a level-$d+1$ LSTM is fed sequentially the information from the children of $v$ at level $d+1$, and produces as output information for $v$ itself. 
Thus, we exploit LSTMs' ability to process sequences of any length to process trees of any branching factor.
The top-level LSTM produces the desired prediction for the target node. 
The architecture requires training $D$ LSTMs, one per tree level.
The LSTMs learn how to summarize the neighborhood up to radius $D$ on the basis of data, avoiding the manual task of synthesizing a fixed set of features.
By dedicating one LSTM to each level, we can tailor the learning (and the LSTM size) to the distance from the target node. 
For instance, in the bipartite graph arising from crowdsourced grading, it is desirable to use different LSTMs for aggregating the edges converging to an item (representing grades received), and for aggregating the edges converting to a user (representing the grades assigned). 

We demonstrate the effectiveness of the proposed approach over four problems. 
The first problem is a synthetic example concerning the crowdsourcing of yes/no labels for items.
The other three are based on real data, and they are the previously mentioned problems of aggregating crowdsourced grades, predicting bitcoin spending, and predicting future reversions of user's edits in Wikipedia.
In all four problems, we show that the ability of MLSL to exploit any feature in the data leads to high performance with minimal feature engineering effort and no apriori model assumptions.
We are making available the open-source code implementing LSTMs and MLSL, along with the datasets, at {\color{blue} \url{https://sites.google.com/view/ml-on-structures}}.

\section{Related Work}

Predicting properties of nodes in graph structures is a common problem that has been widely studied.
Several existing approaches view this as a model-based inference problem.
A model is created, and its parameters are tuned on the basis of the information available; the model is then used to perform inference. 
As the exact probabilistic inference is generally intractable \cite{koller_probabilistic_2009}, most techniques rely on iterative approximation approaches. 
Iterative approximations are also at the root of expectation maximization (EM) \cite{dempster_maximum_1977}.
Iterative parameter estimation has been used, together with Gibbs sampling, to reliably aggregate peer grades in massive on-line courses \cite{piech_tuned_2013}.
Iterative, model-based approaches have also been used for reliably crowdsourcing boolean or multi-class labels \cite{karger_iterative_2011,karger_efficient_2013}.
In these works, a bipartite graph of items and workers is created, and then the worker reliabilities, and item labels or grades, are iteratively estimated until convergence. 

Compared to these models, the benefit of our proposed approach is that it does not require a model, and thus, it can avail itself of all the features that happen to be available. 
For instance, in crowdsourced grading, we can use not only the agreement among the graders to judge their reliability, but also any other information that might be available, such as the time taken to grade, or the time of day, or the number of items previously graded by the user, without need to have a model of how these features might influence grade reliability. 
We will show that this ability can lead to superior performance compared to EM and \cite{karger_iterative_2011} when additional features are available.
On the other hand, machine-learning based approaches such as ours are dependent on the availability of training data, while model-based approaches can be employed even in its absence.

Several approaches have been proposed for summarizing graph structures in feature vectors. 
The algorithm {\em node2vec\/} \cite{grover_node2vec:_2016} enables the construction of feature vectors for graph nodes in such a way that the feature vector optimally represents the node's location in the graph. 
Specifically, the feature vector maximizes the a-posteriori probability of graph neighborhoods given the feature vector. 
The resulting feature vector thus summarizes a node's location in a graph, but it does not summarize the original features of the node, or of its neighbors. 
In contrast, the techniques we introduce allow us to feed to machine learning the node features of an entire graph neighborhood. 

In {\em DeepWalk\/} \cite{perozzi_deepwalk:_2014}, feature vectors for graph nodes are constructed by performing random walks from the nodes, and applying them various summarization techniques to the list of feature vectors of the visited nodes. 
This approach enables the consideration of variable-diameter neighborhoods, in contrast to our exploration, which proceeds strictly breath-first. 
In DeepWalk, the construction of the summarizing feature vector proceeds according to a chosen algorithm, and is not guided by backpropagation from the learning goal. 
In other words, the summarization is not learned from the overall ML task.
In contrast, in our approach the summarization itself, carried out by the LSTMs, is learned via backpropagation from the goal.

LSTMs were proposed to overcome the problem of vanishing gradient over long sequences problems with recurrent neural nets \cite{hochreiter_long_1997,gers_lstm_2001}; they have been widely useful in a wide variety of learning problems; see, e.g., \cite{graves_offline_2009,sundermeyer_lstm_2012}.
Recurrent neural nets and LSTMs have been generalized to multi-dimensional settings \cite{baldi_principled_2003,graves_supervised_2012}. 
The multi-level architecture proposed here can handle arbitrary topologies and non-uniform nodes and edges (as in bipartite graphs), rather than regular n-dimensional lattices, at the cost of exploring smaller neighborhoods around nodes.

Learning over graphs can be reduced to a standard machine-learning problem by summarizing the information available at each node in a fixed set of features. 
This has been done, for instance, with the goal of {\em link prediction,} consisting in predicting which users in a social network will collaborate or connect next \cite{al_hasan_link_2006}.
Graph summarization typically requires deep insight into the problem, in order to design the summary features. 
The multi-level LSTMs we propose here constitute a way of learning such graph summarization. 

Some recent work has looked at the problem of summarizing very large graphs into feature vectors \cite{tang_line:_2015}.
The goals (and methods) are thus different from those in the present paper, where the emphasis consists in considering nodes together with their immediate neighborhoods as input to machine learning.

There is much work on {\em learning with graphs,} where the graph edges encode the similarity between the nodes (rather than features, as in our case); see, e.g., \cite{zhu_semi-supervised_2005,gad_active_2016}. 
This represents an interesting, but orthogonal application to ours.

\section{Learning from Graph Neighborhoods}

We consider a graph $G = (V, E)$ with set of vertices $V$ and edges $E \subs V \times V$. 
We assume that each edge $e \in E$ is labeled with a vector of features $g(e)$ of size $M$.  
Each vertex $v \in V$ is associated with a vector of labels. 
The goal is to learn to predict the vertex labels on the basis of the structure of the graph and the edge labels. 

This setting can model a wide variety of problems. 
Considering only edge features, rather than also vertex features, involves no loss of generality: if there are interesting features associated with the vertices, they can be included in the edges leading to them.
If the goal consists in predicting edge outputs, rather than vertex, one can construct the {\em dual\/} graph $G' = (E, V')$ of $G$, where edges of $G$ are vertices of $G'$, and where $V' = \set{((u,v), (v,w)) \mid (u,v), (v,w) \in E}$. 


\paragraph{Learning method overview.}
Our learning strategy can be summarized as follows.
In order to predict the label of a node $v$, we consider the tree $T_v$ rooted at $v$ and with depth $D$, for some fixed $D > 0$, obtained by unfolding the graph $G$ starting from $v$. 
We then traverse $T_v$ bottom-up, using {\em sequence learners,} defined below, to compute a label for each node from the labels of its children edges and nodes in $T_v$. 
This traversal yields an output label $y_v$ for the root $v$ of the tree. 
In training, the output $y_v$ can be compared with the desired output, a loss be computed, and backpropagated through the tree. 
We now present in detail these steps.

\paragraph{Graph unfolding.}
Given the graph $G = (V, E)$ and a node $v \in V$, along with a depth $D > 0$, we define the {\em full unfolding of $G$ of depth $D$ at $v$} as the tree $T_v$ with root $v$, constructed as follows. 
The root $v$ has depth 0 in $T_v$.  
Each node $u$ of depth $k < D$ in $T_v$ has as children in $T_v$ all nodes $z$ with $(u,z) \in E$; the depth of each such $z$ is one plus the depth of $u$. 
A single graph node may correspond to more than one node in the unfolding.
We will rename the nodes of the unfolding so that they are all distinct; nodes and edges in the unfolding inherit their labels from their correspondents in the graph.

\begin{figure}
\centering
\includegraphics[height=1in, width=3in]{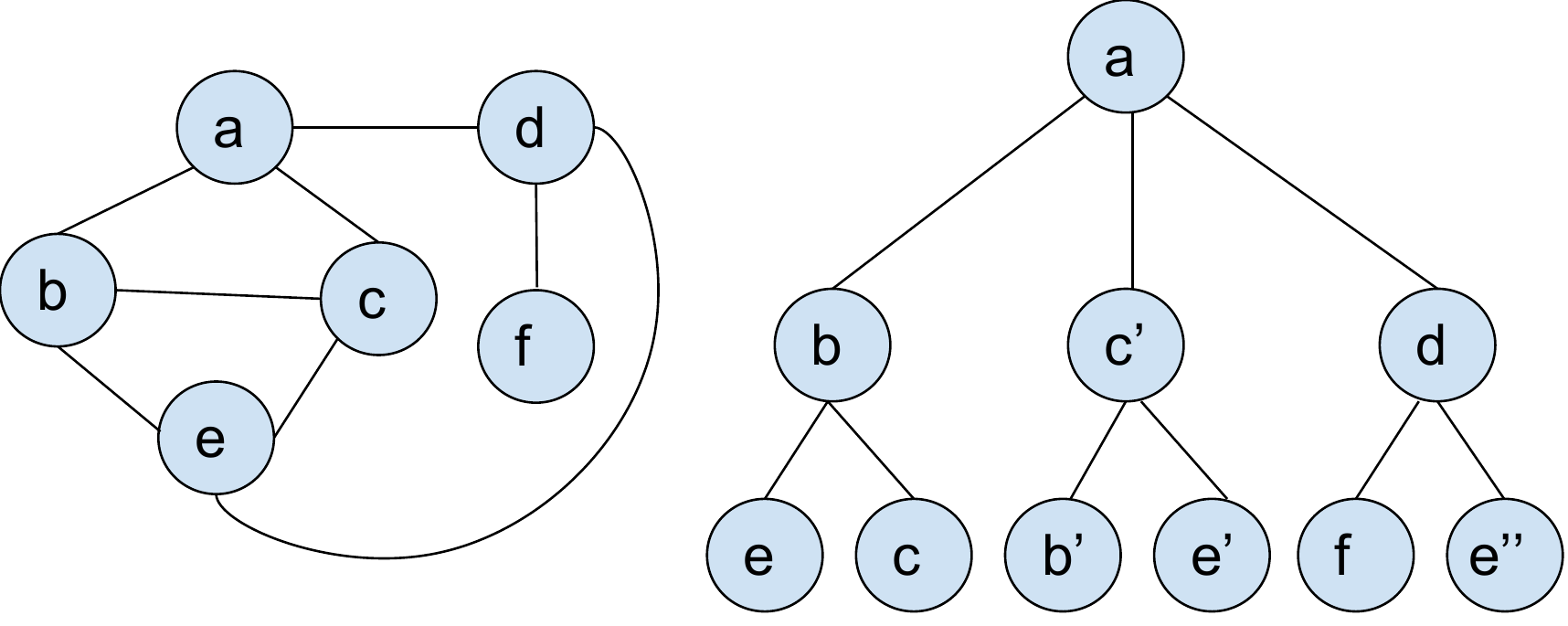}
\caption{An example of a graph and its asymmetric unfolding at node $a$ for depth $2$.
We rename the nodes that appear in many locations so that they have distinct names, for instance, we use $e$, $e'$ and $e''$ to denote the copies of $e$.}
\label{fig-unfolding}
\end{figure}

It is possible to perform learning using {\em asymmetric unfolding,} in which if a node $u$ has parent $u'$, we let the descendants of $u$ be $\set{z \mid (u, z) \in E, z \neq u'}$.
Figure~\ref{fig-unfolding} illustrates a graph and its asymmetric tree unfolding at node $a$ and depth~2.
Which of the two unfolding is more useful depends on the specifics of the learning problem, and we will discuss this choice in our applications.

\paragraph{Sequence learners.}
Our proposed method for learning on graphs leverages {\em sequence learners.}
A sequence learner is a machine-learning algorithm that can accept as input an arbitrary-length sequences of feature vectors, producing a single vector as output. 
{\em Long Short-Term Memory\/} neural nets (LSTMs) \cite{hochreiter_long_1997} are an example of such sequence learners.
We denote a sequence learner parameterized by a vector $w$ of parameters by $L[w]$. 
In LSTMs, the parameter vector $w$ consists of the LSTM weights.
We say that a sequence learner is of shape $(N, K)$ if it accepts a sequence of vectors of size $N$, and produces a vector of size $K$ as output.
We assume that a sequence learner $L[w]$ of shape $(N, K)$ can perform three operations:
\begin{itemize}

\item {\em Forward propagation.} 
Given a input sequence $x^{(1)}, x^{(2)}, \ldots, x^{(n)}$, where each $x^{(i)}$ is a vector of size $N$, compute an output $y$, where $y$ is a vector of size $K$.

\item {\em Loss backpropagation.} 
For a loss function $\loss$, given $\partial \loss / \partial y$ for the output, it can compute $\partial \loss / \partial x^{(j)}$ for each $x^{(1)}, x^{(2)}, \ldots, x^{(N)}$. 
Here, $\partial \loss / \partial y$ is a vector having $\partial \loss / \partial y_i$ as component for each component $y_i$ of $y$, and likewise, $\partial \loss / \partial x^{(j)}$ is a vector with components $\partial \loss / \partial x^{(j)}_k$, for each component $x^{(j)}_k$ of $x^{(j)}$.

\item {\em Parameter update.} 
For a loss function $\loss$, given $\partial \loss / \partial y$ for the output, it can compute a vector $\Delta w$ of parameter updates. 
The parameter updates can be for instance computed via a gradient-descent method, taking $\Delta w = - \alpha \partial \loss / \partial w$ for some $\alpha > 0$, but the precise method varies according to the structure of the sequence learner; see, e.g., \cite{gers_lstm_2001}.

\end{itemize}
In an LSTM, backpropagation and parameter update are performed via {\em backpropagation through time;\/} see \cite{werbos_backpropagation_1990,williams_gradient-based_1995} for details.

\subsection{Multi-Level Sequence Learners}

Given a graph $G$ with labeled edges as above, we now describe the learning architecture, and how to perform the forward step of node label prediction, and the backward step of backpropagation and parameter updates. 
We term our proposed architecture {\em multi-level sequence learners,} or MLSL, for short.\footnote{Open-source code implementing MLSL will be made available by the authors.}

We start by choosing a fixed depth $D > 0$ for the unfolding. 
The prediction and learning is performed via $D$ sequence learners $L_1, L_2, \ldots, L_D$. 
Each sequence learner $L_i$ will be responsible for aggregating information from children at depth $i$ in the unfolding trees, and computing some information for their parent, at depth $i-1$. 
The sequence learner $L_D$ has shape $(M, K_D)$, where $M$ is the size of the edge labels: from the edge labels, it computes a set of features of size $K_D$.  
For each $0 < d < D$, the sequence learner at depth $d$ has shape $(M + K_{d+1}, K_d)$ for some $K_d > 0$, so that it will be able to aggregate the edge labels and the output of the learners below, into a single vector of size $K_d$. 

Note that learners $L_d$ for depth $1 < d \leq D$ can appear multiple times in the tree, once for each node at depth $d-1$ in the tree. 
All of these instances of $L_d$ share the same parameters, but are treated separately in forward and backward propagation.

The behavior of these sequence learners is defined by the parameter vectors $w^{(1)}, \ldots, w^{(D)}$; the goal of the learning is to learn the values for these parameter vectors that minimizes the loss function.
We stress that the sequence learners $L_1, L_2, \ldots, L_D$ and their parameter vectors $w^{(1)}, \ldots, w^{(D)}$ can depend on the depth in the tree (there are $D$ of them, indeed), but they do not depend on the root node $v$ whose label we are trying to predict.

In order to learn, we repeatedly select root nodes $v^* \in V$, for instance looping over them, or via some probability distribution over nodes, and we construct the unfoldings $T_{v^*}$. 
We then perform over $T_{v^*}$ the forward and backpropagation steps, and the parameter update, as follows.

\begin{figure}
\centering
\includegraphics[height=2in, width=3in]{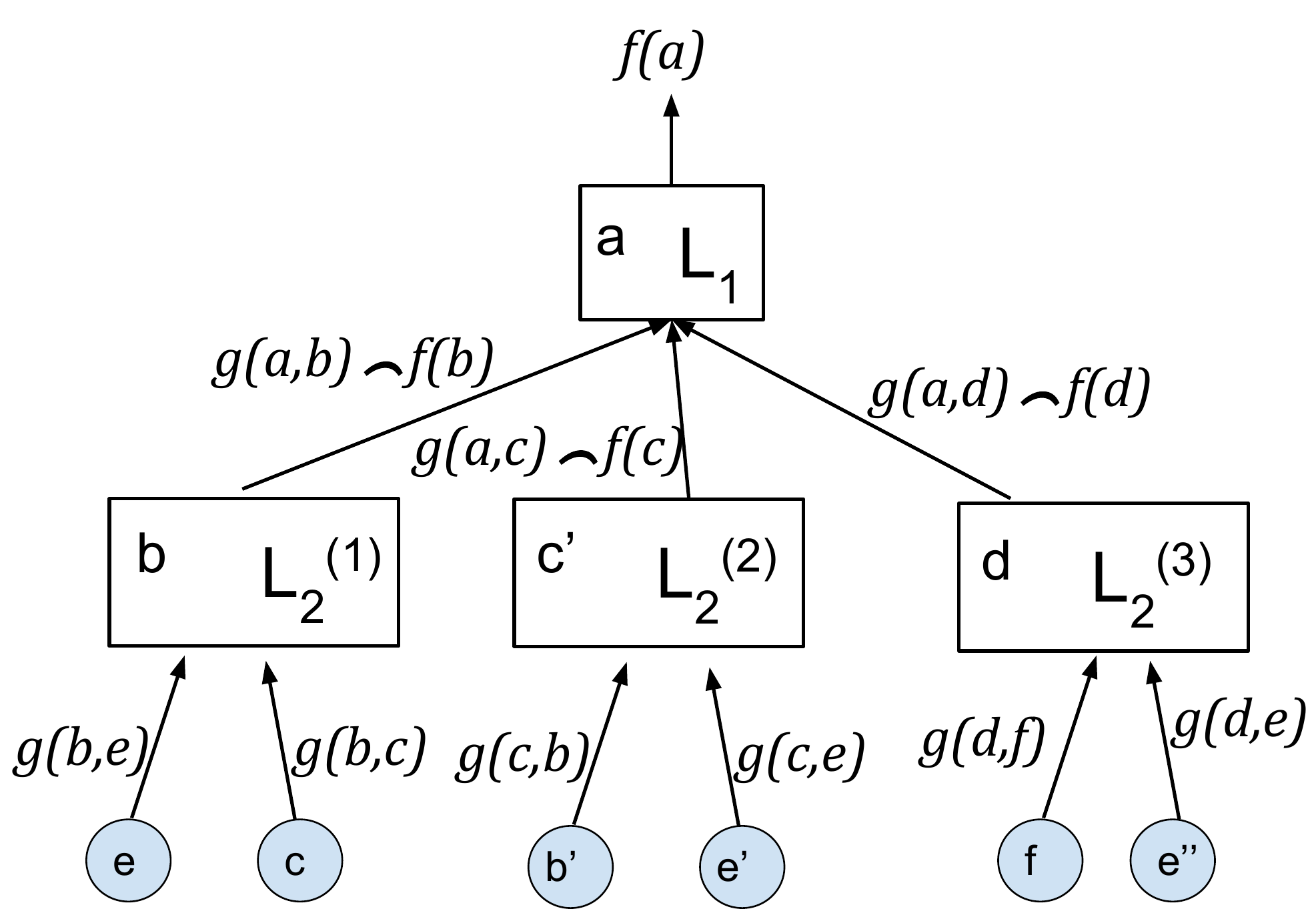}
\caption{Forward propagation corresponding to the tree unfolding of  Figure~\ref{fig-unfolding}. 
The elements of the sequence which is fed to learner $L_1$ consist of the features of the respective edges concatenated with the output from learners below.
Note the use of three instances of the learner $L_2$, one for each depth-2 node in the unfolding. 
These instances share the same parameters.
In the figure, the symbol $\concat$ denotes the concatenation of feature vectors.}
\label{fig-forward}
\end{figure}

\paragraph{Forward propagation.}
The forward propagation step proceeds bottom-up along $T_v$. 
Figure~\ref{fig-forward} illustrates how the sequence learners are applied to an unfolding of the root node $a$ of the graph of Figure~\ref{fig-unfolding} with depth $2$ to yield a prediction for node $a$. 
\begin{itemize}

\item {\em Depth $D$.} 
Consider a node $v$ of depth $D-1$ with children $u_1, \ldots, u_k$ at depth $D$. 
We use the sequence learner $L_D$ to aggregate the sequence of edge labels $g(v, u_1), \ldots, g(v, u_k)$ into a single label $f(v)$ for $v$. 

\item {\em Depth $0 < d < D$.} 
Consider a node $v$ at depth $d-1$ with children $u_1, \ldots, u_k$ at depth $d$. 
We forward to the learner $L_d$ the sequence of vectors $g(v,u_1) \concat f(u_1), \ldots, g(v, u_n) \concat f(u_n)$ obtained by concatenating the feature vectors of the edges from $v$ to the children, with the feature vectors computed by the learners at depth $d+1$. 
The learner $L_d$ will produce a feature vector $f(v)$ for $v$.

\end{itemize}

\paragraph{Backward propagation.}
Once we obtain a vector $y = f(v^*)$ for the root of $T_{v^*}$, we can compute the loss $\loss(y)$, and we can compute $\partial \loss / \partial y$. 
This loss is then backpropagated from the root down to the leaves of $T_{v^*}$, following the topology of the tree (refer again to Figure~\ref{fig-forward}).
Consider a node $v$ at depth $d-1$, for $0 < d \leq D$, with computed feature vector $f(v)$.
We backpropagate through the instance of the learner $L_d$ that computed $f(v)$ the loss, obtaining $\partial \loss / \partial x_i$ for the input vectors $x^{(0)}, \ldots, x^{(k)}$ corresponding to the children $u_1, \ldots, u_k$ of $v$. 
\begin{itemize}

\item If these children are at depth $d < D$, each vector $x^{(j)}$ consists of the concatenation $g(v, u_j) \concat f(u_j) $ of the  features $g(v, u_j)$ from the graph edge, and of the features $f(u_j)$ computed for $u_j$.
As the former require no further backpropagation, we retain the portion $\partial \loss / \partial f(u_j)$ for further backpropagation.

\item At the bottom depth $d = D$ of the tree, each vector $x^{(j)}$ corresponds to the graph edge labels $g(v, u_j)$, and backpropagation terminates.

\end{itemize}

\paragraph{Parameter update (learning).}
Consider a learner $L_d$ for depth $1 \leq d \leq D$, defined by parameters $w^{(d)}$.
To update the parameters $w^{(d)}$, we consider all instances $L_d^{(1)}, \ldots, L_d^{(m)}$ of $L_d$ in the tree $T_{v^*}$, corresponding to the nodes $v_1, \ldots, v_m$ at depth $d$ (refer again to Figure~\ref{fig-forward}). 
For each instance $L_d^{(i)}$, for $i = 1, \ldots, m$, from $\partial \loss / \partial f(v_i)$ we can compute a parameter update $\Delta_i w^{(d)}$.
We can then compute the overall parameter update for $L_d$ as the average
$\Delta w^{(d)} = \bigl(\Delta_1 w^{(d)} + \cdots + \Delta_m w^{(d)} \bigr) / m$ of the updates over the individual instances. 

\paragraph{Preserving learner instance state.}
As mentioned above, a sequence learner for a given depth may occur in several instances in the tree obtained by unfolding the graph (see Figure~\ref{fig-unfolding}). 
Commonly, to perform backpropagation and parameter update though a learner, it is necessary to preserve (or recompute) the state of the learner after the forward propagation step; this is the case, for instance, both for neural nets and for LSTMs. 
Thus, even though all learner instances for depth $d$ are defined by a single parameter vector $w^{(d)}$, it is in general necessary to cache (or reconstruct) the state of every learner instance in the tree individually.

\subsection{Training}

During training, we repeatedly select a target node, unfold the graph, feed the unfolding to the multi-level LSTMs, obtain a prediction, and backpropagate the loss, updating the LSTMs. 
An important choice is the order in which, at each tree node, the edges to children nodes are fed to the LSTM.
The edges can be fed in random order, shuffling the order for every training sample, or they can be fed in some fixed order.
In our applications, we have found each of the two approaches to have uses. 

\section{Applications}

We have implemented multi-level sequence learners on the basis of an LSTM implementation performing backpropagation-though-time learning \cite{graves_supervised_2012}, which we combined with an AdaDelta choice of learning step \cite{zeiler_adadelta:_2012}.
We report the results on one synthetic setting, and three case studies based on real data. 
The code and the datasets can be found at {\color{blue} \url{https://sites.google.com/view/ml-on-structures}}.

For imbalanced datasets, apart from the accuracy (percentage of correct guesses), we report the average recall, which is the unweighted average of the recall of all classes. 
This is suitable in the case of classes of different frequencies, since for highly imbalanced datasets it is easy to inflate the accuracy measure by predicting labels of the most frequent classes.

\subsection{Crowdsourcing boolean labels}

We considered the common boolean crowdsourcing task where users provide yes/no labels for items.
This is modeled as a bipartite graph, with items and users as the two kind of nodes; the edges are labeled with yes/no.
The task consists in reconstructing the most likely labels for the items.
We generated synthetic data similar to the one used in \cite{karger_iterative_2011}. 
In the data, items have a true yes/no label (which is not visible to the inference algorithms), and users have a hidden boolean variable indicating whether they are truthful, or random.
Truthful users report the item label, while random users report yes/no with probability $0.5$ each. 
This is also called the \emph{spammer-hammer} user model.
We report results for a graph of $3000$ users and $3000$ items where item labels are balanced (50\% yes/ 50\% no) and the probability of a user being reliable is 60\%.
Each item gets 3 votes from different users.
We compare three algorithms:
\begin{itemize}
\item The iterative algorithm of \cite{karger_iterative_2011}, abbreviated as KOS.
The algorithm requires no prior.
\item Expectation Maximization (EM) \cite{dempster_maximum_1977}, where user reliability is modeled via a beta distribution.
We used an informative prior (shape parameters $\alpha = 1.2$ and $\beta = 1.0$) for the initial beta distribution which reflects the proportion of reliable users in the graph.
\item Our multi-level sequence learners with depths 1 and 3, denoted 1-MLSL and 3-MLSL, where the output (and memory) sizes of 3-MLSL are $K_2 = K_3 = 3$.
We train on $1,000$ items and test on the remaining $2,000$.
\end{itemize}
For multi-level LSTM, we also consider the case where users have an additional observable feature that is correlated to their truthfulness.
This represents a feature such as ``the user created an account over a week ago'', which is observable, but not part of standard crowdsourcing models.
This feature is true for 90\% of reliable users and for for 40\% of unreliable users.
We denote the algorithms that have access to this extra feature as 1-LSL+ and 3-LSL+; KOS and EM cannot make use of this feature as it is not part of their model.
Our intent is to show how machine-learning approaches such as MLSLs can increase their performance by considering additional features, independently of a model. 

We report the results in Table~\ref{tab:synthetic}.
When no additional information is available, EM is superior to 1-MLSL and slightly superior to 3-MLSL.
When the additional feature is available, both 1-MLSL+ and 3-MLSL+ learn its usefulness, and perform best.

\begin{table}
\begin{center}
\begin{tabular}{ | c | c |}
\hline
  Method & Accuracy\\
  \hline
  KOS & 0.8016 \\
  EM   & 0.9136 \\ \hline
\end{tabular}
\quad
\begin{tabular}{ | c | c |}
\hline
 Method & Accuracy \\
  \hline
1-MLSL &  0.8945 \\
3-MLSL & 0.9045 \\
1-MLSL+ & 0.9565\\
3-MLSL+ & 0.9650\\ \hline
\end{tabular}
\caption{Performance of KOS \cite{karger_iterative_2011}, EM (Expectation Maximization) and multi-level sequence learners (MLSLs) of different depths.}
\label{tab:synthetic}
\end{center}
\end{table}

\subsection{Peer Grading}

We considered a dataset containing peer grading data from computer science classes. 
The data comes from an online tool that lets students submit homework and grade each other's submissions. 
Each sumission is typically reviewed by 3 to 6 other students.
The data is a bipartite graph of users and submissions, as in the previous crowdsourcing application.
Users assign grades to items in a predefined range (in our case, all grades are normalized in the 0-10 range).
Each edge is labeled with the grade, and with some additional features: the time when the student started grading the submission, and the time when they submitted the grade.
We treat this as a classification task, where the classes are the integer grades 0, 1, \ldots, 10; the ground truth is provided by instructor grades, available on a subset of submissions.
Our dataset contined 1,773 labeled (instructor-graded) submissions; we used 1,500 for training and 273 for testing.

We compare three methods.
One is simple average of provided grades, rounded to the closest integer.
Another method is based on expectation maximization (EM), iteratively learning the accuracy of users and estimating the grades.
Finally, we employed MLSL with the following features (derived from the graph): the time to complete a review, the amount of time between review completion and review deadline, and the median grade received by the student in the assignment. The output of the learner at level $2$ is of size $3$ where it reaches its peak for this experiment.

\begin{table}
\begin{center}
    \begin{tabular}{| c| c | c |}
    \hline
     Method & Accuracy & \pbox{20 cm}{Average Recall}  \\ \hline
     Average & 0.5432 & 0.3316 \\
     EM-based & 0.5662 &  0.3591 \\ 
     1-MLSL &0.6044    & 0.3897   \\ 
     2-MLSL &0.6010  &  0.3913\\
   	\hline
    \end{tabular}
    \caption {Performance of EM and 1,2-depth MLSL on peer grading data.}
    \label{tab:crowdgrader}
    \end{center}
\end{table}

Table~\ref{tab:crowdgrader} shows the results. The 1- and 2-depth MLSL methods are superior to both the EM-based approach and average.
Average recall appears low due to the very high class imbalance of the dataset: some low homework grades are very rare, and mistakes in these rare grades have high impact.

\subsection{Prediction of Wikipedia Reversions}

Wikipedia is a popular crowdsourced knowledge repository with contributions from people all around the world and in various languages.
Users occasionally add contributions that are reverted by other users, either due to their low quality, or as part of a quarrel, or simply due to carelessness. 
Our interest is in predicting, for each user, whether the user's next edit will be reverted. 
We note that this is a different (and harder) question than the question of whether a specific edit, whose features are already known, will be reverted in the future \cite{adler_wikipedia_2011}.

We model the user interactions in Wikipedia as a multi-graph with users as nodes.
An edge $e$ from $u_2$ to $u_1$ represents a ``implicit interaction'' of users $u_2$ and $u_1$, occurring when $u_2$ creates a revision $r_2$ immediately following a revision $r_1$ by $u_1$.
Such an edge $e$ is labeled with a feature vector consisting of the edit distances $d(r_1,r_2)$, $d(r_0,r_2)$ and $d(r_0, r_1)$, where $r_0$ is the revision immediately preceding $r_1$, and $d(\cdot)$ is edit distance.
The feature vector contains also the elapsed times between the revisions, and the {\em quality\/} of $r_1$ measured from $r_2$, defined by 
$d(r_0,r_1) / \bigl(d(r_0,r_2) - d(r_1,r_2)\bigr)$
\cite{adler_content-driven_2007}.

Since the English Wikipedia has a very large dataset, for this experiment we used the complete dumps of the Asturian Wikipedia (Asturian is a language in Spain). 
The graph consists of over $32,000$ nodes (users) and over $45,000$ edge (edits among users). 
To obtain the labels for each user, we consider the state of this graph at a time 30 days before the last date of content available in the dump; this leaves ample time for reversions to occur in the extra 30 days, ensuring that we label users correctly.
To train the model, we repeatedly pick an edit by a user, and we construct the graph neighborhood around the user consisting only of the edits {\em preceding\/} the selected edit (we want to predict the future on the basis of the past).
We label the user with yes/no, according to whether the selected edit was reverted, or not. 
This local neighborhood graph is then fed to the MLSL.
We performed training on 60\% of the data and validated with the remaining 40\%. 
We trained over 30 models for each depth and validated them by measuring the average recall and F1-scores for both labels. 
Table~\ref{wiki-table} shows the average results for each depth level. 
We observe that F-1 scores for both ``reversion'' and ``no reversion'' labels were high. 
Moreover, these results show improvement in performance for increasing depth. 

\begin{table}[]
	\centering
	\begin{tabular}{|l|l|l|l|}
  \hline
& Average & F-1 & F-1 \\
& Recall & reverted & not reverted \\ \hline 
	  1-MLSL & 0.8468 & 0.8204 & 0.8798 \\ 
		2-MLSL & 0.8485 & 0.8259 & 0.8817 \\ 
		3-MLSL & 0.8508 & 0.8288 & 0.8836 \\ \hline
	\end{tabular}
	\caption{Prediction of reversions in the Asturian Wikipedia, using MLSL of depths 1, 2, 3.}
	\label{wiki-table}
\end{table}

\subsection{Prediction of Bitcoin Spending}

The blockchain is the public immutable distributed ledger where Bitcoin transactions are recorded \cite{nakamoto_bitcoin:_2008}. 
In Bitcoin, coins are held by {\em addresses,} which are hash values; these address identifiers are used by their owners to anonymously hold bitcoins, with ownership provable with public key cryptography.
A Bitcoin transaction involves a set of source addresses, and a set of destination addresses: all coins in the source addresses are gathered, and they are then sent in various amounts to the destination addresses.

Mining data on the blockchain is challenging \cite{meiklejohn_fistful_2013} due to the anonymity of addresses.
We use data from the blockchain to predict whether an address will spend the funds that were deposited to it.

We obtain a dataset of addresses by using a slice of the blockchain.
In particular, we consider all the addresses where deposits happened in a short range of 101 blocks, from 200,000 to 200,100 (included) .
They contain 15,709 unique addresses where deposits took place.
Looking at the state of the blockchain after 50,000 blocks (which corresponds to roughly one year later as each block is mined on average every 10 minutes), 3,717 of those addresses still had funds sitting: we call these ``hoarding addresses''.
The goal is to predict which addresses are hoarding addresses, and which spent the funds.
We randomly split the 15,709 addresses into a training set of 10,000 and a validation set of 5,709 addresses.


\begin{table}
    \begin{tabular}{| l | l | l | l | l |}
    \hline
     &Accuracy & \pbox{20 cm}{Avg. \\ Recall} & \pbox{20cm}{ F-1\\ `spent'} & \pbox{20cm}{ F-1 \\`hoard'} \\ \hline
    Baseline & 0.6325 & 0.4944 & 0.7586 & 0.2303\\ 
    1-MLSL &  0.7533 & 0.7881 & 0.8172 & 0.6206 \\ 
    2-MLSL & 0.7826 &  0.7901 & 0.8450 & 0.6361\\ 
    3-MLSL & 0.7731 & 0.7837 & 0.8367 & 0.6284\\ 
    \hline
    \end{tabular}
    \caption {The prediction results on blockchain addresses using baseline approach, and MLSL of depths 1, 2, 3.}
    \label{tab:blockchain}
\end{table}

We built a graph with addresses as nodes, and transactions as edges. 
Each edge was labeled with features of the transaction: its time, amount of funds transmitted, number of recipients, and so forth, for a total of 9 features.
We compared two different algorithms:
\begin{itemize}
\item Baseline: an informative guess; it guesses a label with a probability equal to its percentage in the training set.
\item MLSL of depths 1, 2, 3.  
The outputs and memory sizes of the learners for the reported results are $K_2 = K_3 = 3$. 
Increasing these to $5$ maintained virtually the same performance while increasing training time. 
Using only $1$ output and memory cell was not providing any advances in performance.

\end{itemize}

Table~\ref{tab:blockchain} shows the results.  
Using the baseline we get poor results; the F-1 score for the smaller class (the `hoarding' addresses) is particularly low. 
Tapping the transaction history and using only one level the learner already provides a good prediction and an average recall approaching 80\%.
Increasing the number of levels from 1 to 2 enhances the quality of the prediction as it digests more information from the history of transactions.
Increasing the levels beyond 2 does not lead to better results, with this dataset.

\subsection{Discussion}

The results from the above applications show that MLSL can provide good predictive performance over a wide variety of problems, without need for devising application-tailored models. 
If sufficient training data is available, MLSL can use the graph representation of the problem and any available features to achieve high performance.

One of our conclusions is that the order of processing the nodes during training matters.
In crowdsourced grading, randomly shuffling the order of edges for a learning instance as it is used in different iterations during the training process, was superior to using a fixed order.
For Bitcoin, on the other hand, feeding edges in temporal order worked best. This seems intuitive, as the transactions happened in some temporal order.

One challenge was the choice of learning rates for the various levels.
As the gradient backpropagates across the multiple levels of LSTMs, it becomes progressively smaller. 
To successfully learn we needed to use different learning rates for the LSTMs at different levels, as the top levels will tend to learn faster.

\bibliography{multi_layer_learning_on_graphs}
\bibliographystyle{plain}

\end{document}